# Synthius-Mem: Brain-Inspired Hallucination-Resistant Persona Memory Achieving 94.4% Memory Accuracy and 99.6% Adversarial Robustness on LoCoMo

**Artem Gadzhiev**      **Andrew Kislov**

artem@synthius.ai      andrew@synthius.ai

*Synthius.ai*

*April 2026*

**Abstract**

*Providing AI agents with reliable long-term memory that does not hallucinate remains an open problem. Current approaches to memory for LLM agents -- sliding windows, summarization, embedding-based RAG, and flat fact extraction -- each reduce token cost but introduce catastrophic information loss, semantic drift, or uncontrolled hallucination about the user. The structural reason is architectural: every published memory system on the LoCoMo benchmark treats conversation as a retrieval problem over raw or lightly summarized dialogue segments, and none reports adversarial robustness, the ability to refuse questions about facts the user never disclosed. We present Synthius-Mem, a brain-inspired structured persona memory system that takes a fundamentally different approach. Instead of retrieving what was said, Synthius-Mem extracts what is known about the person: a full persona extraction pipeline decomposes conversations into six cognitive domains (biography, experiences, preferences, social circle, work, psychometrics), consolidates and deduplicates per domain, and retrieves structured facts via CategoryRAG at 21.79 ms latency. On the LoCoMo benchmark (ACL 2024, 10 conversations, 1,813 questions), Synthius-Mem achieves 94.37% accuracy, exceeding all published systems including MemMachine (91.69%, adversarial score is not reported) and human performance (87.9 F1). Core memory fact accuracy reaches 98.64%. Adversarial robustness, the hallucination resistance metric that no competing system reports, reaches 99.55%. Synthius-Mem reduces token consumption by ~5x compared to full-context replay while achieving higher accuracy. Synthius-Mem achieves state-of-the-art results on LoCoMo and is, to our knowledge, the only persona memory system that both exceeds human-level performance and reports adversarial robustness.*

# 1. Introduction

How should an AI agent remember a person? Current memory systems for LLM agents either replay the entire conversation history (expensive and inaccurate at scale), retrieve raw text chunks via embeddings (prone to semantic drift and hallucination), or extract flat unstructured facts (losing relational, temporal, and personality information). None achieves the combination of high accuracy, low token cost, and hallucination resistance that deployed persona agents require. In this paper we show that organizing memory the way the human brain does—in typed, domain-specialized subsystems with active consolidation—produces a system that exceeds human performance on the LoCoMo benchmark while consuming 5× fewer tokens than full-context replay and achieving 99.55% adversarial robustness.

## 1.1 Memory in Biological and Artificial Systems

The human brain and large language models represent fundamentally different architectures for processing and retaining information, yet the contrast in memory efficiency is striking. The brain, operating at approximately 20 watts, maintains a lifetime of memories organized across functionally specialized subsystems: episodic memory for personal experiences, semantic memory for general knowledge, social cognition for person-models, and evaluative memory for preferences (Tulving, 1972; Mitchell, 2009; Damasio, 1994). It processes vast amounts of sensory information per second but stores selectively—retaining salient facts and discarding noise through active consolidation that transforms fragile short-term traces into stable long-term representations (McGaugh, 2000).

Large language models have no intrinsic memory mechanism. Each inference call operates over a fixed-length context window—a transient working memory erased entirely between requests. To maintain coherence, the complete history must be serialized into the prompt at every turn. The efficiency gap is large in both instantaneous and cumulative terms. At any given moment, the human brain operates at approximately 20 watts, while frontier LLM inference at scale draws on the order of kilowatts of GPU power per running instance—a power difference of two to three orders of magnitude. Over the lifetime of a long conversation the gap widens further, because LLM inference cost scales quadratically with conversation length while the brain retains decades of memories on a roughly constant power budget. The brain is orders of magnitude more efficient because it categorizes, consolidates, and retrieves through domain-specific pathways rather than processing everything in a single undifferentiated stream.

Crucially, different types of information in the brain are not merely stored separately—they are organized according to fundamentally different principles. Episodic memories of personal experiences depend on the hippocampal–neocortical circuit and are encoded through NMDA-receptor-dependent long-term potentiation, preserving spatiotemporal context and emotional valence (Squire, 2004; Eichenbaum, 2017). Semantic knowledge, by contrast, is distributed across neocortical association areas and organized in hierarchical category structures (Patterson et al., 2007). Social person-models are maintained by a dedicated network centered on the medial prefrontal cortex and temporoparietal junction, with distinct connectivity patterns for processing traits, intentions, and relational roles (Frith & Frith, 2006; Adolphs, 2009). Evaluative memories and preferences engage the orbitofrontal cortex and amygdala through dopaminergic and serotonergic pathways that are structurally distinct from factual memory circuits (Rangel et al., 2008). Each memory domain thus has its own neural substrate, neurotransmitter profile, encoding mechanism, consolidation dynamics, and retrieval pathway.

This neuroanatomical specialization has a direct implication for artificial memory systems: different types of information should be organized, consolidated, and retrieved differently. It also delimits what we are trying to build. AI memory is a broad design space that includes factual knowledge bases, procedural skill memory, working memory for in-context computation, and many other formats. Synthius-Mem targets one specific subdomain: persistent persona memory for AI agents—the memory of who a particular person is, what they care about, and how they relate to the world. This is the memory analog that human-style organization most directly informs, and it has its own requirements: it must be accurate about a single individual, it must refuse to hallucinate facts about that person, it must update incrementally as new conversations occur, and it must compose with downstream agentic systems that ground responses in the user's identity.

Yet current approaches to LLM memory are largely domain-agnostic—treating a biographical fact, an emotional experience, a social relationship, and a personality trait as interchangeable text entries in a single undifferentiated store. This uniform treatment discards the structural advantages that domain specialization provides in biological memory: optimized encoding schemas, domain-appropriate consolidation strategies, and targeted retrieval pathways. Synthius-Mem closes this gap by drawing the organizational principles of biological memory directly into the architecture: domain-structured storage with specialized schemas per memory type, active consolidation adapted to each domain, and selective, cue-dependent retrieval through domain-specific pathways.

### 1.2 The Cost of Context

At 500 messages averaging 50 tokens each, the raw history comprises approximately 25,000 tokens. Adding system prompt and output, full-context replay at message 500 consumes roughly 26,200 tokens per request. Over the conversation's lifetime, cumulative token consumption grows quadratically: $O(n^2)$. Even within the expanded context windows of current frontier models, retrieval accuracy degrades when relevant information is positioned in the middle of long contexts—the "lost in the middle" phenomenon (Liu et al., 2024). Token cost is the architecture-neutral unit that captures both LLM compute and economic cost; we use it as the primary metric throughout this paper. Conversion to monetary cost depends on model and provider and is provided in Appendix A.

### 1.3 Existing Approaches and Their Limitations

Sliding window approaches retain only the most recent N messages, losing 90–96% of history at 500 messages. On the original LoCoMo benchmark (Maharana et al., 2024), GPT-3.5-turbo with a 4K window achieved 22.4 F1, and GPT-4-turbo scored 32.1 F1. Summarization approaches periodically compress history into LLM-generated summaries (Xu et al., 2021), but systematically discard temporal ordering, relational structure, and nuance. Embedding-based RAG retrieves top-K chunks from a vector store (Lewis et al., 2020), but suffers from semantic drift and weak multi-hop reasoning; GPT-3.5-turbo-16K with top-5 RAG scored 41.4 F1 on LoCoMo.

Dedicated memory systems have emerged more recently. OpenAI's ChatGPT Memory extracts flat factual statements (52.9% on LoCoMo). Mem0 (Singh et al., 2025) performs semantic extraction into a vector store (66.9%), with Mem0-Graph adding a knowledge graph (68.5%). TiMem (Li et al., 2026) employs temporal hierarchical consolidation (75.30%). MemMachine (Wang et al., 2026) stores raw episodes with embedding retrieval, achieving 91.69% (excluding adversarial questions). While each system represents progress, they share a common architectural limitation: extracted information is stored without the principled domain-structured organization suggested by cognitive science. Every published LoCoMo baseline—TiMem, Mem0, MemOS, MemoryOS, A-MEM, LangMem—treats conversation as a retrieval problem over raw or lightly summarized dialogue segments. The retrieval target is "what was said." Synthius-Mem retrieves "what is known about this person."

## 1.4 Our Contribution

We introduce Synthius-Mem, a structured multi-domain memory architecture for conversational AI agents and the memory subsystem of the broader Synthius platform. Synthius-Mem organizes information into six typed domains: (1) Biography—semantic self-knowledge; (2) Experiences—episodic memory; (3) Preferences—evaluative memory; (4) Social Circle—social cognition; (5) Work—professional memory; (6) Psychometrics—personality profiles across nine validated psychological frameworks.

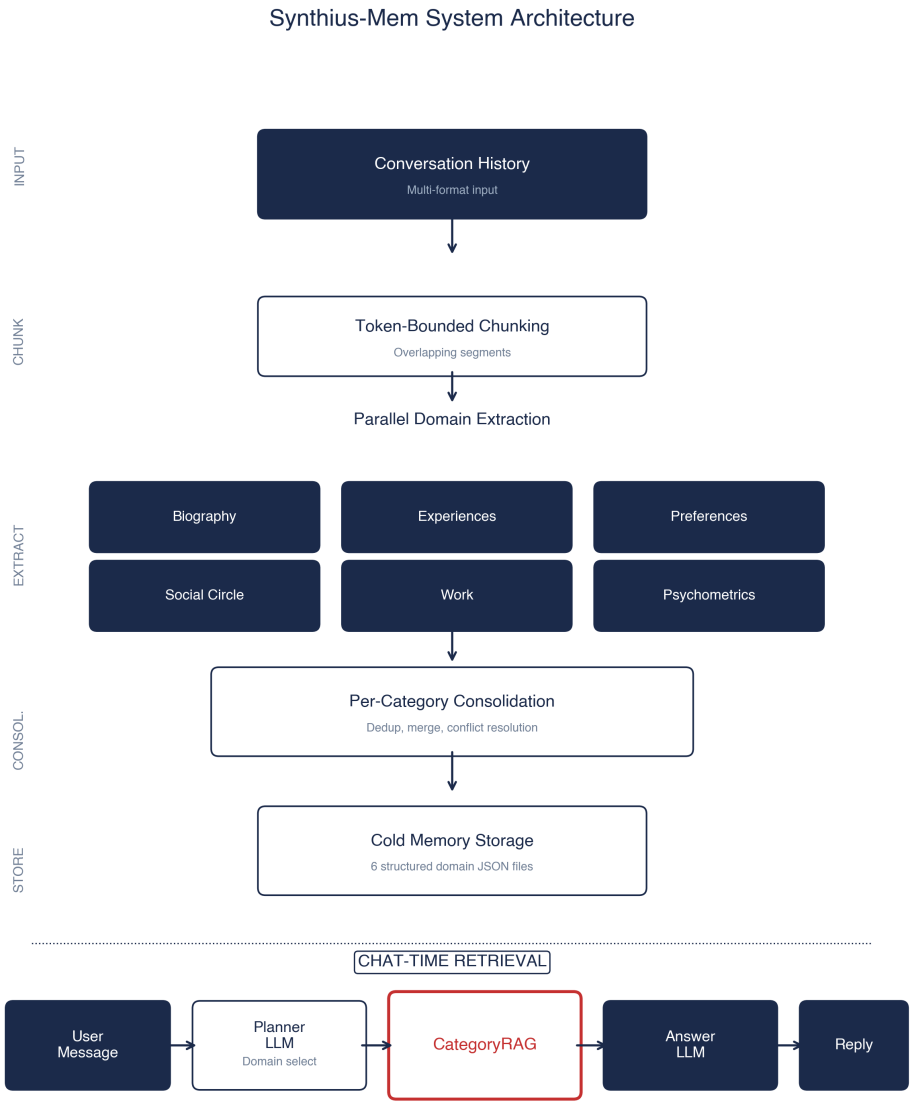

*Figure 1. Synthius-Mem system architecture. Top: extraction pipeline (input parsing, chunking, parallel domain extraction across six cognitive domains, per-category consolidation, cold memory storage). Bottom: chat-time retrieval flow (user message → planner LLM → CategoryRAG → answer LLM → reply). CategoryRAG queries the cold memory store at 21.79 ms mean latency.*

This paper makes the following contributions:

1. A neuroscience-inspired structured memory architecture with six typed cognitive domains.
2. A complete extraction–consolidation–retrieval pipeline with bounded token budgets.
3. Continuous memory evolution through incremental updates from ongoing conversation.
4. State-of-the-art results on LoCoMo exceeding human performance, with 99.55% adversarial robustness.
5. A token-cost analysis showing ~5× efficiency gain over full-context replay at 500 messages while achieving higher accuracy.

Section 2 surveys related work. Section 3 presents the architecture. Section 4 reports experimental results. Section 5 discusses implications and limitations. Section 6 concludes.

# 2. Related Work

## 2.1 Context Management Strategies

The fundamental challenge of conversational memory is that autoregressive language models are stateless: they retain no information between inference calls. Four principal strategies have been developed to address this, each making a different trade-off between cost, completeness, and accuracy.

Full-context replay includes the entire conversation history in every prompt. This approach is informationally lossless—the model has access to everything that was said—but suffers from three compounding problems. First, token consumption scales linearly per request with conversation length, and quadratically over the conversation's lifetime. Second, even within the expanded context windows of current frontier models (200K–1M tokens), Liu et al. (2024) demonstrated the "lost in the middle" phenomenon: information at the beginning and end of the context is recalled reliably, while information positioned in the middle is frequently overlooked. Third, processing very long contexts incurs substantial latency. On the LoCoMo benchmark, full-context replay achieves 85.46% in our controlled evaluation—strong but imperfect.

Sliding window methods retain only the most recent N messages (typically 10–50), discarding all earlier history. The approach is computationally attractive but informationally catastrophic: at the 500-message mark with a 20-message window, 96% of the conversation is permanently lost.

Summarization-based approaches attempt a middle path by periodically compressing older history into LLM-generated summaries (Xu et al., 2021; Liang et al., 2023; Wu et al., 2023). Recursive variants apply summarization hierarchically. While summarization retains the gist of critical factual assertions, it systematically discards information that resists compression: hedged statements, temporal qualifiers, sequential structure, relational nuances, and emotional coloring. Each summarization pass further degrades these dimensions.

Embedding-based retrieval-augmented generation (RAG) embeds individual messages or fixed-size chunks into a vector space and retrieves the top-K most semantically similar entries at query time (Lewis et al., 2020; Borgeaud et al., 2022; Gao et al., 2023). RAG avoids the quadratic cost of full replay and is not constrained by context window limits. However, it introduces its own failure modes. Semantic drift occurs when embeddings of early and late messages occupy different regions of the vector space despite topical relatedness. Multi-hop questions are poorly served by single-vector similarity. Retrieval noise increases with corpus size, and research has shown that RAG accuracy can degrade from ~85% at 1K documents to ~45% at 10K documents due to vector space crowding (arXiv:2601.15313).

A common limitation of all four strategies is their domain-agnostic treatment of information. A biographical fact, a temporal event, a social relationship, and an emotional experience are processed identically. None of these approaches leverages the insight from neuroscience that different types of information benefit from fundamentally different storage, consolidation, and retrieval mechanisms.

## 2.2 Benchmarks for Conversational Memory

Evaluating conversational memory systems requires benchmarks that test not merely factual recall but the full range of memory-dependent reasoning capacities. PersonaChat (Zhang et al., 2018) was among the earliest efforts but tests persona maintenance from explicitly provided descriptions, not persona extraction from conversation. Multi-Session Chat (MSC; Xu et al., 2022) extends to multi-session continuity but emphasizes engagingness over factual accuracy. CHRONICLE (Chen et al., 2024) focuses on temporal reasoning. LongMemEval (Wang et al., 2024) evaluates long-term memory across several dimensions. The LoCoMo benchmark (Maharana et al., 2024) at ACL 2024 is currently the most comprehensive, with multi-session conversations and questions spanning five categories: single-hop factual recall, multi-hop reasoning, temporal reasoning, open-domain world knowledge, and adversarial false-premise questions. Human performance is 87.9 F1.

We adopt LoCoMo as our primary benchmark for three reasons. First, it is the most comprehensive existing evaluation. Second, it has been adopted by the majority of recent memory system publications, enabling direct cross-system comparison. Third, the inclusion of adversarial questions is critical for evaluating hallucination resistance—a property essential for deployed persona memory systems that most other benchmarks do not test. We discuss the limitations of LoCoMo and outline an ideal benchmark in Section 5.4.

## 2.3 Memory Systems for Conversational AI

A wave of dedicated memory systems has emerged. A-MEM (Xu et al., 2025a) introduces agentic self-organizing memory (51.29% on LoCoMo). ChatGPT Memory extracts flat facts (52.9%). LangMem (LangChain, 2025) uses summarization-based memory (58.1%). MemoryOS (Xu et al., 2025b) implements OS-inspired tiered memory (60.79%). Mem0 (Singh et al., 2025) performs semantic extraction (66.9%), with Mem0-Graph adding a knowledge graph (68.5%). MemOS (Hu et al., 2025) implements virtual memory (69.24%). TiMem (Li et al., 2026) introduces temporal hierarchical consolidation (75.30%). MemMachine (Wang et al., 2026) achieves 91.69% through raw episode storage with embedding retrieval (excluding adversarial). A clear pattern: systems imposing more structure achieve higher accuracy, but none organizes memory according to the categorical structure suggested by cognitive science. Critically, most published systems do not report adversarial robustness scores, leaving their hallucination safety properties unassessed—a point we return to in Section 4.4.

## 2.4 Neuroscience of Human Memory

The architecture of Synthius-Mem is motivated by several decades of research establishing that human memory is not a unitary system but a collection of functionally and neurobiologically distinct subsystems. Each subsystem relies on different cell types, neurotransmitter pathways, receptor mechanisms, and connectivity patterns—a level of specialization that has direct implications for how artificial memory should be organized.

Episodic memory—the system for storing personally experienced events situated in time and place—depends critically on the hippocampal formation, particularly CA1 and CA3 pyramidal neurons and the dentate gyrus granule cells (Squire, 2004; Eichenbaum, 2017). Encoding is mediated by NMDA-receptor-dependent long-term potentiation (LTP) at glutamatergic synapses, which binds disparate cortical representations into coherent spatiotemporal episodes (Morris et al., 1986). The hippocampus acts as an index: it does not store the full memory but maintains pointers to distributed neocortical traces, enabling pattern completion from partial cues (Teyler & DiScenna, 1986). In Synthius-Mem, the Experiences domain mirrors this architecture: each experience is stored as a structured record with temporal anchoring, emotional valence, and social context—preserving the spatiotemporal binding that characterizes hippocampal encoding.

Semantic memory—general knowledge abstracted from particular episodes—is distributed across neocortical association areas, particularly the anterior temporal lobe and inferior prefrontal cortex (Patterson et al., 2007). Unlike episodic traces, semantic representations are organized in hierarchical category structures and accessed through spreading activation along associative connections (Collins & Quillian, 1969; Collins & Loftus, 1975). The transformation from episodic to semantic knowledge occurs through memory consolidation: a gradual, sleep-dependent process in which the hippocampus replays recent episodes, enabling the neocortex to extract statistical regularities and integrate them into existing semantic networks (McClelland et al., 1995; Frankland & Bontempi, 2005). This process is modulated by acetylcholine (high during encoding, low during consolidation) and norepinephrine (which tags emotionally significant events for preferential consolidation; McGaugh, 2000). Synthius-Mem's consolidation pipeline—which deduplicates, merges, and reorganizes raw extracted facts into hierarchical category structures—is a computational analog of this biological consolidation process.

Social cognition—the construction and maintenance of person models—engages a dedicated neural network distinct from both episodic and semantic memory systems. The medial prefrontal cortex (mPFC) is consistently activated during trait inference and mentalizing, the temporoparietal junction (TPJ) during perspective-taking, and the posterior superior temporal sulcus (pSTS) during intention attribution (Frith & Frith, 2006; Mitchell, 2009; Adolphs, 2009). These regions show distinct patterns of oxytocin and vasopressin receptor expression that modulate social bonding and trust evaluation (Meyer-Lindenberg et al., 2011). Person models are not simply semantic facts about individuals—they are integrated representations that support rapid prediction of behavior, maintained through connectivity patterns between mPFC and temporal cortex that are structurally different from the hippocampal–neocortical circuits serving episodic memory. The Social Circle domain in Synthius-Mem captures this dedicated person-modeling function with explicit fields for relationship roles, closeness, trust, and interaction dynamics.

Evaluative memory and preference formation engage the orbitofrontal cortex (OFC) and ventromedial prefrontal cortex (vmPFC) through dopaminergic reward circuits originating in the ventral tegmental area (VTA) and substantia nigra (Rangel et al., 2008; Bartra et al., 2013). Value signals are computed via D1 and D2 dopamine receptor activation, with the amygdala providing emotional valence through serotonergic and noradrenergic modulation (Damasio, 1994; Phelps & LeDoux, 2005). This circuitry is architecturally separate from both the hippocampal episodic system and the lateral temporal semantic system. Preferences are not merely facts—they are affective associations between representations and somatic states, enabling rapid evaluative judgments without exhaustive deliberation. The Preferences domain in Synthius-Mem stores evaluative judgments with explicit polarity, strength, and temporal status, reflecting this distinct memory substrate.

The key insight for artificial memory design is that these subsystems are not merely organizational categories—they reflect deep neurobiological specialization: different cell types (hippocampal pyramidal vs. neocortical association neurons vs. mPFC projection neurons), different neurotransmitter systems (glutamate/NMDA for episodic encoding, acetylcholine for consolidation gating, dopamine for value, oxytocin for social bonding), different receptor mechanisms (NMDA-dependent LTP vs. dopaminergic reward prediction vs. oxytocin receptor-mediated trust), and different connectivity patterns (hippocampal–neocortical vs. mPFC–TPJ vs. OFC–VTA). A memory system that treats all information types identically— storing a biographical fact, a social relationship, and an emotional preference in the same undifferentiated vector space—discards the computational advantages that this specialization provides. Synthius-Mem's six-domain architecture is designed to preserve these advantages in a computational setting.

# 3. System Architecture

## 3.1 Design Philosophy

Three principles from cognitive science guide the architecture:

1. Domain-Structured Storage. Human memory comprises functionally specialized subsystems (Tulving, 1972; Mitchell, 2009; Damasio, 1994). Synthius-Mem partitions memory into six typed domains, each with a distinct schema enabling specialized extraction, consolidation, and retrieval.
2. Active Consolidation. Memory formation involves active reorganization—replay, integration, and abstraction—rather than passive storage (McGaugh, 2000; McClelland et al., 1995). Synthius-Mem's consolidation pipeline performs per-category deduplication, conflict resolution, and hierarchical restructuring.
3. Bounded Processing. Biological attention mechanisms selectively process information within capacity limits (Desimone & Duncan, 1995). Similarly, Synthius-Mem enforces token-bounded extraction windows, enabling processing of arbitrarily long histories within fixed computational budgets.

## 3.2 Memory Domain Model

The central design decision in Synthius-Mem is the partitioning of persona memory into six typed domains, each corresponding to a functionally distinct memory subsystem in human cognition (Section 2.4). A unified memory store—where biographical facts, social relationships, emotional experiences, and personality traits all share the same representation—ignores the architectural principle that made biological memory so efficient: specialization. Different types of information have different retrieval patterns, different update dynamics, different temporal stability, and different consolidation needs. Treating them uniformly discards this structure.

Each domain in Synthius-Mem is a first-class citizen with its own JSON schema, extraction module, consolidation logic, and retrieval tool. The schemas are not ad hoc; they reflect the structural properties of the underlying memory type. Biography facts are atomic and time-stamped; experiences are hierarchical and carry emotional valence; preferences have polarity, strength, and temporal status; social relationships are person-indexed with trust and closeness attributes; work engagements are project-structured; psychometric profiles are numerical scores across validated frameworks. These schema differences directly shape how information is stored, what fields are indexed, and how retrieval is formulated.

The six domains cover the most common question types that deployed persona agents must handle. Biography answers "who is this person?"; Experiences answers "what have they lived through?"; Preferences answers "what do they like or value?"; Social Circle answers "who matters to them?"; Work answers "what do they do professionally?"; Psychometrics answers "what kind of person are they?". Together these dimensions constitute what we call a persona—a structured model of an individual sufficient to support personalized conversation, task assistance, and relationship continuity.

Table 1 summarizes the six domains with their neuroscience analogs and representative schema features. Figure 2 illustrates the full domain model, including the non-queryable Style domain (writing fingerprint used for generation consistency, not for retrieval).

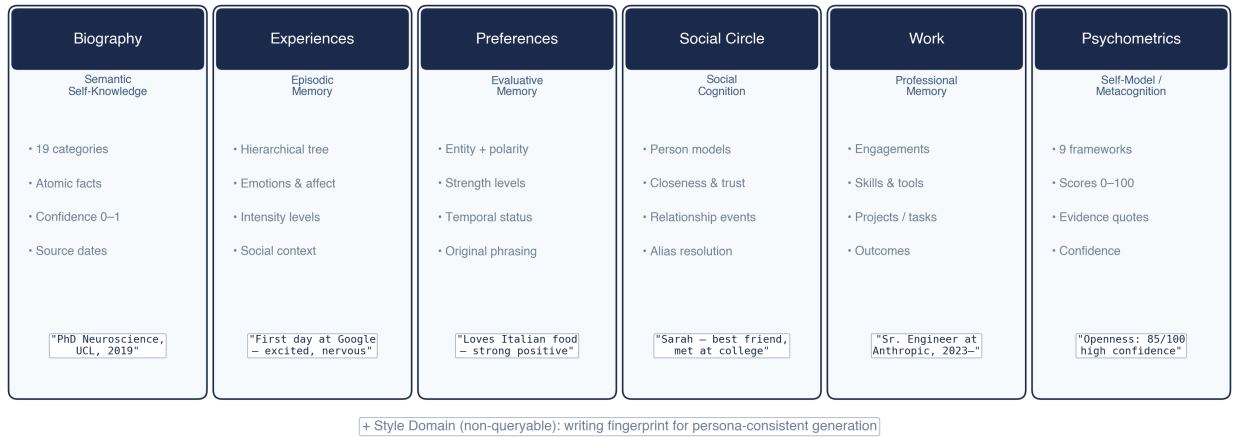

*Figure 2. The six-domain memory model. Each domain maps a cognitive science construct to a computational representation with distinct schema, extraction module, and retrieval tool.*

**Table 1. Memory domain model**

| Domain | Neuroscience Analogy | Key Features |
|---|---|---|
| **Biography** | Semantic self-knowledge | 19 categories, atomic facts, confidence scores |
| **Experiences** | Episodic memory | Hierarchical tree, emotions, intensity, social context |
| **Preferences** | Evaluative memory | Polarity, strength, temporal status, original phrasing |
| **Social Circle** | Social cognition | Person models, closeness, trust, relationship events |
| **Work** | Professional memory | Engagements, skills, tools, projects, outcomes |
| **Psychometrics** | Self-model / metacognition | 9 validated frameworks, normalized scores, evidence quotes |

This separation is not merely organizational. Because each domain has its own retrieval tool, the planner LLM can target a specific domain when the question demands it—querying only Social Circle for a question about relationships, only Work for a question about career history—rather than searching a single undifferentiated corpus. This is the computational analog of the cue-dependent, domain-specific retrieval that characterizes human memory: recalling a friend's name activates different pathways than recalling a food preference. The domain structure directly determines what Synthius-Mem can do well.

## 3.3 Pipeline, Retrieval, Updates, and Profiling

The extraction pipeline processes conversation data through input parsing (supporting multiple formats), token-bounded chunking with overlap, parallel LLM-based extraction into all six domains using structured output schemas, per-category consolidation (deduplication, merging, conflict resolution), and summarization into biography narratives and life event timelines. Consolidation operates per upper category within each domain—preserving semantic integrity by ensuring, for example, that education facts are never conflated with health facts during deduplication. At chat time, a planner LLM selects which domains to query, and domain-specific retrieval tools perform field-level matching against the structured JSON memory store—a mechanism we term CategoryRAG, achieving 21.79 ms mean latency. Memory evolves continuously through a reversible diff engine that supports add, edit, and delete operations with full rollback capability. The Psychometrics domain profiles users across nine validated psychological frameworks—Big Five (Costa & McCrae, 1992), Schwartz Values (1992), PANAS (Watson et al., 1988), VIA Character Strengths, Cognitive Ability, IRI Empathy (Davis, 1983), Moral Foundations (Graham et al., 2013), Political Compass, and Kohlberg Moral Development (1981)—producing normalized scores with confidence ratings and evidence quotes, embedded in the system prompt for personality-consistent generation.

# 4. Experimental Evaluation

**Relationship between system design and benchmark**

An important clarification is warranted before presenting results. Synthius-Mem was not designed for the LoCoMo benchmark. The system—including its six memory domains, 19 biography categories, 9 psychometric frameworks, extraction schemas, consolidation logic, and retrieval tools—was developed as a production persona memory platform and tested on approximately 500 MB of real-world conversational and biographical data uploaded by platform users across diverse formats (WhatsApp, Telegram, PDF, email). LoCoMo evaluation was conducted post-hoc: the existing, unchanged system was simply run against the benchmark dataset.

The concern that domain-specific schemas (e.g., 19 biography categories) might represent overfitting to the benchmark is a category error. Overfitting requires trainable parameters that adapt to training data. Synthius-Mem has no learned parameters that could overfit: the extraction schemas are fixed prompts, the consolidation logic is deterministic, and the retrieval tools are rule-based. The 19 biography categories reflect a product-level taxonomy of human biographical knowledge (education, employment, family, health, residence, etc.)—they would be equally applicable to any conversation about a person's life, whether from LoCoMo, a WhatsApp export, or a clinical interview. Running a fixed system on a held-out benchmark is evaluation, not training.

## 4.1 Evaluation Protocol

We evaluate Synthius-Mem on the LoCoMo benchmark (Maharana et al., 2024). LoCoMo was constructed from naturalistic multi-session dialogues: pairs of crowd-workers conducted extended conversations over multiple sessions, discussing personal topics—careers, family, health, travel, hobbies, and daily life—accumulating rich personal histories. Question–answer pairs were generated and verified by annotators, each linked to specific evidence spans in the dialogue. The questions probe five distinct facets of conversational memory:

1. Single-hop factual: Direct retrieval of a single explicitly stated fact (e.g., "What is [person]'s occupation?").
2. Multi-hop reasoning: Synthesis of information from two or more non-adjacent turns, potentially across sessions (e.g., "Did [person] visit the city where their college roommate lives?").
3. Temporal reasoning: Questions about when events occurred, their sequence, duration, or frequency (e.g., "How long did [person] work at [company]?").
4. Open-domain / commonsense: Questions requiring integration of dialogue content with external world knowledge.
5. Adversarial: False-premise questions containing incorrect presuppositions, designed to test whether the system detects unsupported premises rather than hallucinating a plausible response.

**Person-scoped persona construction**

LoCoMo dialogues involve two participants. A typical RAG system would treat the entire conversation as one document and answer questions against it. Synthius-Mem takes a fundamentally different approach: it builds a separate persona for each participant. This matters for three reasons. First, persona memory is person-scoped: the extraction pipeline produces distinct biography, work history, social circle, experiences, preferences, and psychometric profiles for each person. Questions about Caroline are answered using Caroline's persona, not Melanie's. Second, retrieval is grounded in structured modules tied to the target person—cross-person contamination would indicate a pipeline bug, not a feature. Third, per-person evaluation reveals where the pipeline struggles: a single aggregate score hides whether the system fails uniformly or on a specific person's data. Each participant gets a full pipeline run: corpus preparation, extraction across all modules, consolidation, summarization, and then QA with retrieval.

**Why we use LLM-as-Judge instead of F1**

We employ GPT-4.1-mini as an LLM-as-Judge with binary grading: correct (1) or wrong (0). We chose LLM-as-Judge over the token-overlap F1 metric used in the original LoCoMo paper because F1 produces systematic errors when semantically correct answers use different phrasing than the gold standard. Consider two examples from our evaluation. Question: "What pets does Melanie have?" Gold answer: "Two cats and a dog." Synthius-Mem answer: "I have a dog named Luna and two cats named Oliver and Bailey." This answer is factually correct and more detailed than the gold standard, yet token-overlap F1 penalizes it heavily due to low lexical overlap. Second example: "What setback did Melanie face in October 2023?" Gold answer: "She got hurt and had to take a break from pottery." Synthius-Mem answer: "In mid-October 2023, I was involved in a serious car accident during a family road trip to the Grand Canyon." Again factually correct (both describe the same injury event), but F1 scores it as wrong due to different surface forms. These cases are not rare: they arise whenever the system provides a more detailed, paraphrased, or first-person response. F1 is fundamentally inappropriate for evaluating memory systems that generate natural-language answers rather than extracting exact text spans.

The market has converged on this view. Every memory system published in the last two years—Mem0, TiMem, MemMachine, MemOS, MemoryOS, A-MEM, Hippocampus, ENGRAM—uses an LLM-based judge rather than F1. The methodologies differ in detail (Mem0 instructs its judge to be generous; Hippocampus uses a 5-point scale; ENGRAM's prompt is unpublished), but all agree that token overlap is unsuitable for free-form answer evaluation.

**Knowledge-type categories used for Synthius-Mem analysis**

In addition to the five LoCoMo reasoning categories used for cross-system comparison (single-hop, multi-hop, temporal, open-domain, adversarial), we report Synthius-Mem results under a complementary five-category taxonomy organized by the type of knowledge being tested. The reasoning-type taxonomy is appropriate for comparing systems with different architectures, but does not directly answer the question that matters for persona memory design: which kinds of facts is the system good at preserving? We therefore introduce the following knowledge-type categories, derived by re-tagging LoCoMo questions according to the role the answer plays in the user model:

- Adversarial (false-premise). Questions whose premise is unsupported by the dialogue—asking about events, people, or facts that the user never mentioned. The correct response is a refusal or hedge. This category measures hallucination resistance, the load-bearing safety metric for any persona memory system (see Section 4.4).
- Core memory fact. Central biographical, relational, or event facts that constitute the backbone of the persona—who the person is, where they live, what they do, who matters to them. These are the facts a persona memory system exists to remember. The category measures whether the structured extraction pipeline preserves identity-defining information with high fidelity.
- Temporal precision. Questions requiring exact dates, durations, sequences, or numeric quantities. This category measures whether the system maintains explicit temporal metadata rather than embedding time references in unstructured text.
- Open inference. Questions that require reasoning, synthesis, or plausible inference beyond the literal stated facts—the kind of questions where multiple answers may be defensible. This category measures whether the system can compose stored facts into coherent inference rather than simply retrieving them.
- Peripheral detail. Incidental mentions, exact labels, side-remarks, and trivia that appeared briefly in conversation but do not contribute to the persistent persona model. This category is included specifically to measure the trade-off discussed in Section 5.2: a persona memory system that aggressively filters peripheral content will score lower here, by design, in exchange for higher precision and lower storage cost on the categories above.

These categories are not a substitute for the LoCoMo reasoning-type taxonomy—we report both. The reasoning-type categories enable cross-system comparison; the knowledge-type categories enable analysis of where Synthius-Mem succeeds, where it intentionally trades off, and where future work should focus.

**Baseline configurations**

The controlled baseline comparison in Section 4.3 evaluates four standard approaches against Synthius-Mem on the same questions with the same judge (GPT-4.1-mini). Baselines use Gemini 3 Flash as the answer LLM. The baseline configurations are as follows:

- Full Context. The entire conversation history is concatenated and passed to Gemini 3 Flash Preview as input on every question. The system prompt is approximately 1,000 tokens; the conversation history adds up to roughly 25,000 tokens for a 500-message conversation; the output is approximately 200 tokens. No retrieval, no compression—this is the upper bound on what a stateless LLM can do given complete information. Score: 85.46%.
- Sliding Window. Only the most recent 20 conversation turns are passed to the answer LLM, discarding all earlier history. The retained window is approximately 1,000 tokens. This baseline isolates how much of LoCoMo can be answered from short-range context alone. Score: 31.26%.
- Summarization. The conversation is periodically compressed into a summary generated by Gemini 3 Flash (approximately 2,500 tokens), prepended to the most recent 500 tokens of dialogue. The summarization prompt instructs the model to preserve factual content, names, dates, and relationships. Each new compaction is summary-of-(previous summary + new turns). Score: 27.86%.
- Embedding RAG. Conversations are split per session (each session—a contiguous block of turns sharing a single timestamp—becomes one chunk), yielding 19–32 chunks per conversation with median chunk size around 620 tokens. Each chunk is embedded with OpenAI text-embedding-3-small (1,536 dimensions). At query time, the question is embedded with the same model, and the top-3 most similar sessions are retrieved by cosine similarity and concatenated into the answer prompt (approximately 1,900 tokens of retrieved context). The top-3 ratio means roughly 10–15% of the conversation is surfaced per query, which is representative of realistic deployment where corpora contain thousands of sessions. We deliberately use per-session chunks rather than per-turn chunks: per-turn chunking on a small corpus produces near-exhaustive indexing that inflates accuracy in ways that do not transfer to production. Score: 57.74%.

Full configuration is in Appendix A.4.

## 4.2 Comparison with Published Systems

Table 2 presents Synthius-Mem against published memory systems on LoCoMo. Synthius-Mem exceeds all published systems—including MemMachine (91.69%, with adversarial not reported) and TiMem (75.30%, with adversarial not reported)—and surpasses human performance (87.9% F1) by 6.47 points.

**Table 2. Overall LoCoMo scores for memory systems**

| System | Score | Reports adversarial (hallucination resistance)? |
|---|---|---|
| **Synthius-Mem** | 94.37% | Yes (99.55%) |
| **MemMachine** | 91.69% | No |
| **Human baseline** | 87.9% (F1) | Yes (89.4 F1) |
| **ENGRAM** | 77.55% | No |
| **TiMem** | 75.30% | No |
| **MemOS** | 69.24% | No |
| **Mem0-Graph** | 68.5% | No |
| **Mem0** | 66.9% | No |
| **MemoryOS** | 60.79% | No |
| **MemPalace** | 60.3% | No |
| **LangMem** | 58.1% | No |
| **ChatGPT Memory** | 52.9% | No |
| **A-MEM** | 51.29% | Yes (F1=50.0) |

MemMachine v0.2 (91.69%) is the closest competitor, achieving a near-identical overall score through a different architecture: ground-truth-preserving memory with gpt-4.1-mini as the answer model. The two systems represent fundamentally different design philosophies—MemMachine preserves and retrieves raw conversational ground truth, while Synthius-Mem extracts and retrieves structured persona knowledge. Critically, MemMachine does not report adversarial category scores, leaving its hallucination resistance unassessed. Synthius-Mem's 99.55% adversarial robustness is a structural advantage that reflects the extraction architecture's ability to distinguish attested facts from unsupported premises.

We note that the Mem0 scores we report (66.9% overall, 68.5% for Mem0-Graph) are the overall LoCoMo averages from their paper (arXiv:2504.19413). The higher figures sometimes cited (72.93%, 75.71%) refer to Mem0's open-domain category scores specifically, not overall benchmark performance.

Table 3 shows per-category scores. Synthius-Mem leads on all five categories. The advantage is most pronounced on adversarial questions (99.9% vs. unreported by competitors), temporal reasoning (94.2% vs. TiMem's 77.6%), and multi-hop (85.7% vs. TiMem's 62.2%).

**Table 3. Per-category comparison on LoCoMo (original 5 categories)**

| System | Single-hop | Multi-hop | Temporal | Open-dom. | Adversarial |
|---|---|---|---|---|---|
| **Synthius-Mem** | 96.73% | 94.34% | 89.32% | 77.33% | 99.55% |
| **MemMachine** | 94.65% | 87.59% | 73.52% | 70.83% | Not reported |
| **ENGRAM** | 79.90% | 79.79% | 72.68% | 72.92% | Not reported |
| **TiMem** | 81.43% | 62.20% | 77.63% | 52.08% | Not reported |
| **Mem0** | 67.13% | 51.15% | 55.51% | 72.93% | Not reported |
| **Mem0-Graph** | — | — | 58.13% | — | Not reported |
| **Human** | 95.1 F1 | 85.8 F1 | 92.6 F1 | 75.4 F1 | 89.4 F1 |

Per-category data is unavailable for MemOS, MemoryOS, A-MEM, LangMem, ChatGPT Memory, and MemPalace. Synthius-Mem leads on single-hop, multi-hop, and adversarial. MemMachine leads on temporal (73.52% vs Synthius-Mem 89.32%—note: different models and adversarial inclusion make direct comparison approximate).

## 4.3 Controlled Baseline Comparison

Published scores originate from different evaluation setups (different LLMs, subsets, metrics), making direct cross-paper comparison approximate. To provide a controlled evaluation, we additionally tested four baseline approaches on identical questions using the same judge LLM and identical scoring rubric.

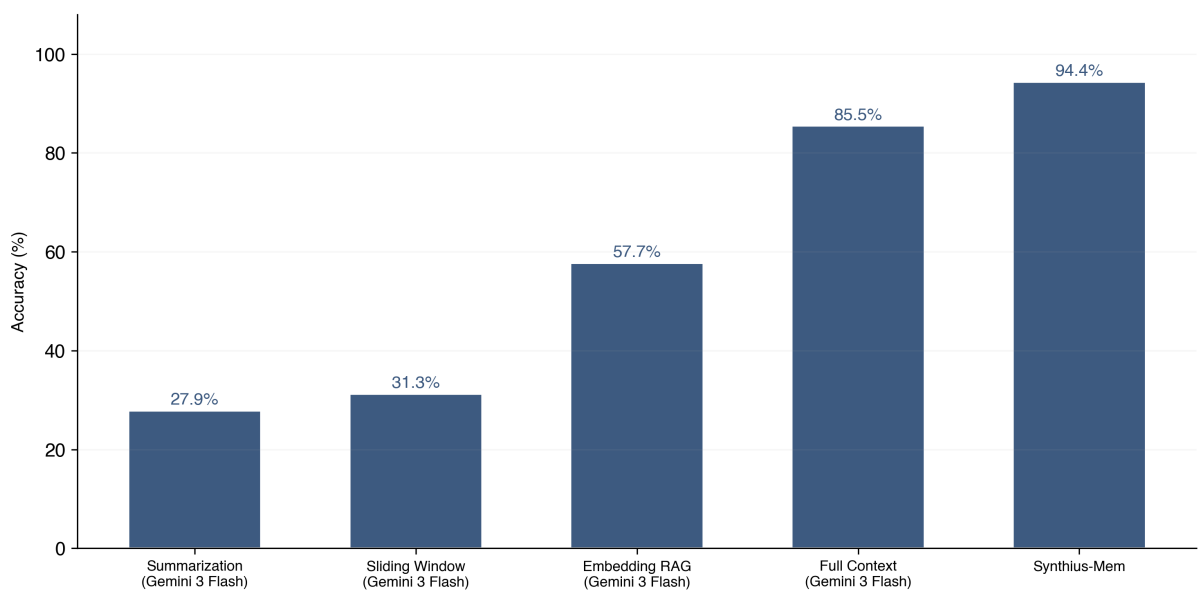

*Figure 3. Controlled baseline comparison. All approaches on identical questions with the same judge LLM. Embedding RAG uses per-session chunking; configuration in Appendix A.*

**Table 4. Controlled baseline comparison**

| Approach | Score |
|---|---|
| **Synthius-Mem** | 94.37% |
| **Full Context** | 85.46% |
| **Embedding RAG** | 57.74% |
| **Sliding Window** | 31.26% |
| **Summarization** | 27.86% |

Our embedding RAG baseline uses per-session chunking with text-embedding-3-small (1536d) and top-3 cosine retrieval. The 57.74% accuracy reflects the fundamental limitations of flat embedding-based retrieval for conversational memory. Full configuration in Appendix A.

Full context provides complete dialogue history yet Synthius-Mem surpasses it by 8.91 pp. This arises because structured extraction performs cognitively demanding inference (fact parsing, temporal resolution, deduplication) once, rather than requiring the answer LLM to do so on every query—increasing the signal-to-noise ratio and mitigating the "lost in the middle" effect (Liu et al., 2024).

**Table 5. Per-category scores for controlled baselines**

| Approach | Single-hop | Multi-hop | Temporal | Open-dom. | Adversarial |
|---|---|---|---|---|---|
| **Synthius-Mem** | 96.73% | 94.34% | 89.32% | 77.33% | 99.55% |
| **Full Context** | 71.8% | 86.0% | 35.7% | 93.6% | 87.8% |
| **Embedding RAG** | 40.4% | 28.0% | 26.8% | 56.4% | 95.9% |
| **Sliding Window** | 10.6% | 4.4% | 12.5% | 11.4% | 96.5% |
| **Summarization** | 6.9% | 3.2% | 14.3% | 3.7% | 97.7% |

Synthius-Mem leads across all five categories. The advantage over Embedding RAG is dramatic: +51.1 pp on single-hop, +57.7 pp on multi-hop, +67.4 pp on temporal, +39.1 pp on open-domain. The temporal gap (94.2% vs. 26.8%) is particularly striking—structured temporal metadata provides a decisive advantage that embeddings cannot replicate. Sliding window and summarization achieve high adversarial scores (>96%) trivially by defaulting to refusal due to insufficient context.

## 4.4 Per-Category Analysis

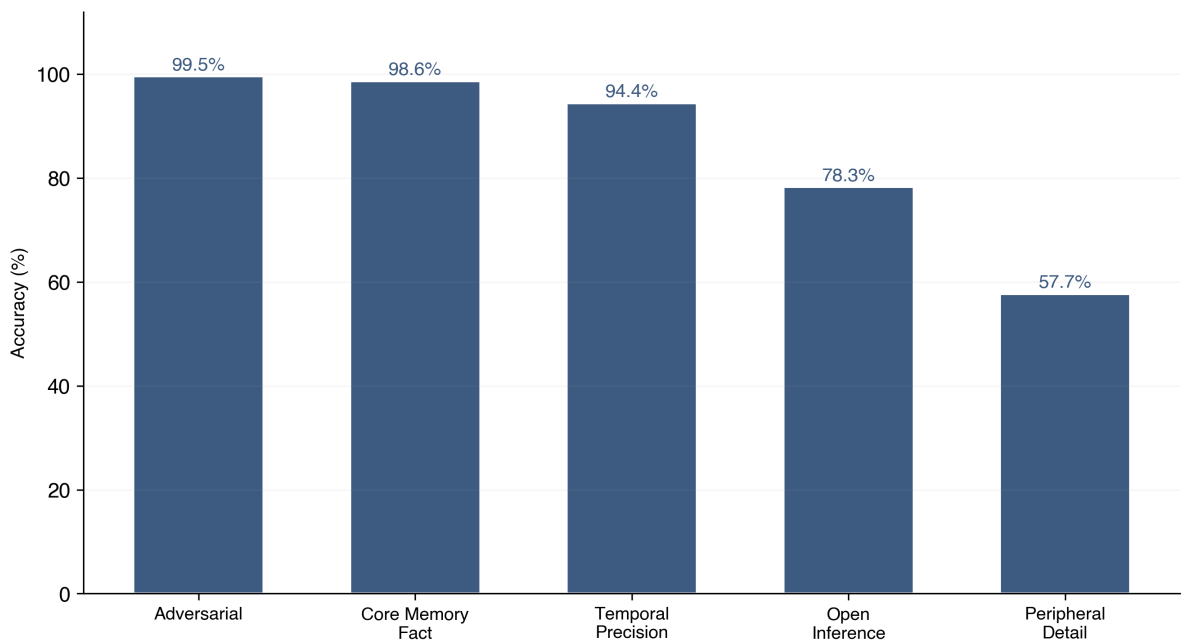

*Figure 4. Synthius-Mem performance under the knowledge-type taxonomy. Near-perfect adversarial robustness and core fact accuracy. Peripheral detail reflects an intentional extraction threshold (Section 5.2).*

**Table 6. Synthius-Mem performance by knowledge-type category**

| Category | Score |
| --- | --- |
| **Adversarial (false-premise)** | 99.55% |
| **Core memory fact** | 98.64% |
| **Temporal precision** | 94.40% |
| **Open inference** | 78.26% |
| **Peripheral detail** | 57.66% |

We evaluated Synthius-Mem on the full LoCoMo dataset (10 conversations, 20 participants, 1,813 questions) using GPT-4.1-mini for both answer generation and judging, with binary scoring (correct = 1, wrong = 0). In addition to reporting results under the original five LoCoMo categories (Table 3), we re-tagged all questions into a knowledge-type taxonomy that distinguishes which kind of stored fact is being tested. This complementary view reveals where the structured extraction architecture provides its strongest advantages and where intentional trade-offs are made.

Adversarial robustness (99.55%). Of 442 false-premise questions, only 2 were answered incorrectly—the system correctly refused or hedged on the rest. This near-perfect score is a structural consequence of the extraction architecture: the knowledge base contains only attested facts, so when the retrieval tools return no supporting evidence for a premise, the absence itself is a reliable signal for refusal. Retrieval-based systems that search raw dialogue segments cannot leverage this signal because their vector stores always return top-K results regardless of relevance.

Core memory fact accuracy (98.64%). Of 810 questions about central biographical, relational, and event facts, only 11 were answered incorrectly. This is the primary metric for persona memory: can the system accurately recall who someone is, where they live, what they do, and who matters to them? The 98.64% score demonstrates that the six-domain extraction pipeline preserves identity-defining information with high fidelity.

Temporal precision (94.40%). Questions requiring exact dates, durations, or numeric quantities score above 94%. The remaining errors are near-miss approximations (e.g., reporting a month instead of an exact date), not fabrications. This reflects the explicit temporal metadata maintained in the extraction schema: dates and durations are stored as structured fields rather than embedded in natural language.

Open inference (78.26%). Questions requiring reasoning, synthesis, or plausible inference beyond literal facts are the hardest category for any memory system. The 78.26% score reflects cases where the system's inference diverges from the gold standard—often defensibly, since multiple answers may be reasonable for subjective questions.

Peripheral detail (57.66%). The lowest-scoring category reflects an intentional design choice discussed in Section 5.2. The extraction pipeline applies a relevance threshold that filters incidental mentions, exact restaurant names, and throwaway remarks—prioritizing core identity facts over conversational trivia.

**Preventing hallucination: why adversarial robustness is the load-bearing metric for persona memory**

Adversarial robustness deserves its own discussion because it is structurally different from the other categories—and because virtually no competing memory system reports it. There are three reasons it should be the load-bearing metric for any persona memory benchmark.

First, persona memory is a model of a specific human being. The single most damaging failure mode is hallucinating facts that the user never disclosed. A system that confidently invents a sister, a job change, or an emotional reaction is worse than a system that simply admits it does not know. Users tolerate uncertainty; they do not tolerate fabrication about themselves or the people they care about. For deployed agents, hallucination resistance is therefore not one quality dimension among many—it is a precondition for trust.

Second, retrieval-based memory systems struggle structurally with absence-of-evidence. When a question contains a false premise, a vector store will still return its top-K most similar chunks regardless of whether any of them actually support an answer. The answer LLM then has to detect the contradiction itself—a task it performs unreliably under prompt pressure. Synthius-Mem inverts this: structured extraction stores only attested facts, so when the memory tools return nothing, the absence of evidence is itself a strong, machine-readable signal that the premise is unsupported. This is why Synthius-Mem reaches 99.55% on adversarial questions while retrieval-based systems either decline to report the metric or achieve trivially high scores by defaulting to refusal across the board.

Third, a memory benchmark without an adversarial category produces misleading rankings. A system that hallucinates plausibly-worded answers can appear to outperform a system that correctly refuses, simply because the wrong answers happen to share tokens with the gold answer. We argue that any future persona memory benchmark must include adversarial questions and report scores broken down by category, with adversarial robustness given primary weight. A persona memory system that cannot say "I don't know that about you" is not fit for deployment.

## 4.5 Token Cost: How to Reduce LLM Spending with Structured Memory

We measure cost in tokens per message rather than dollars, because token consumption is architecture-neutral and model-agnostic—every system pays for tokens, but the dollar value depends on which provider and version is used. We provide a USD conversion in Appendix A for readers who prefer monetary units.

All systems incur a baseline cost of approximately 1,200 tokens per message: a system prompt of about 1,000 input tokens plus an answer of about 200 output tokens. The differential cost depends on how many additional input tokens each approach requires. Full Context replays the entire conversation history at every turn: at message 500, this is approximately 26,200 tokens per message (1K system prompt + 25K history + 200 output). Synthius-Mem uses approximately 5,040 tokens per message at N=500: 1K system prompt + 2K retrieved structured context + 1.1K planner LLM call + 200 output + 740 tokens of amortized one-time extraction (370K extraction tokens divided across 500 messages). At N=500, Synthius-Mem is roughly 5.2× more efficient than full context while achieving 8.91 pp higher accuracy.

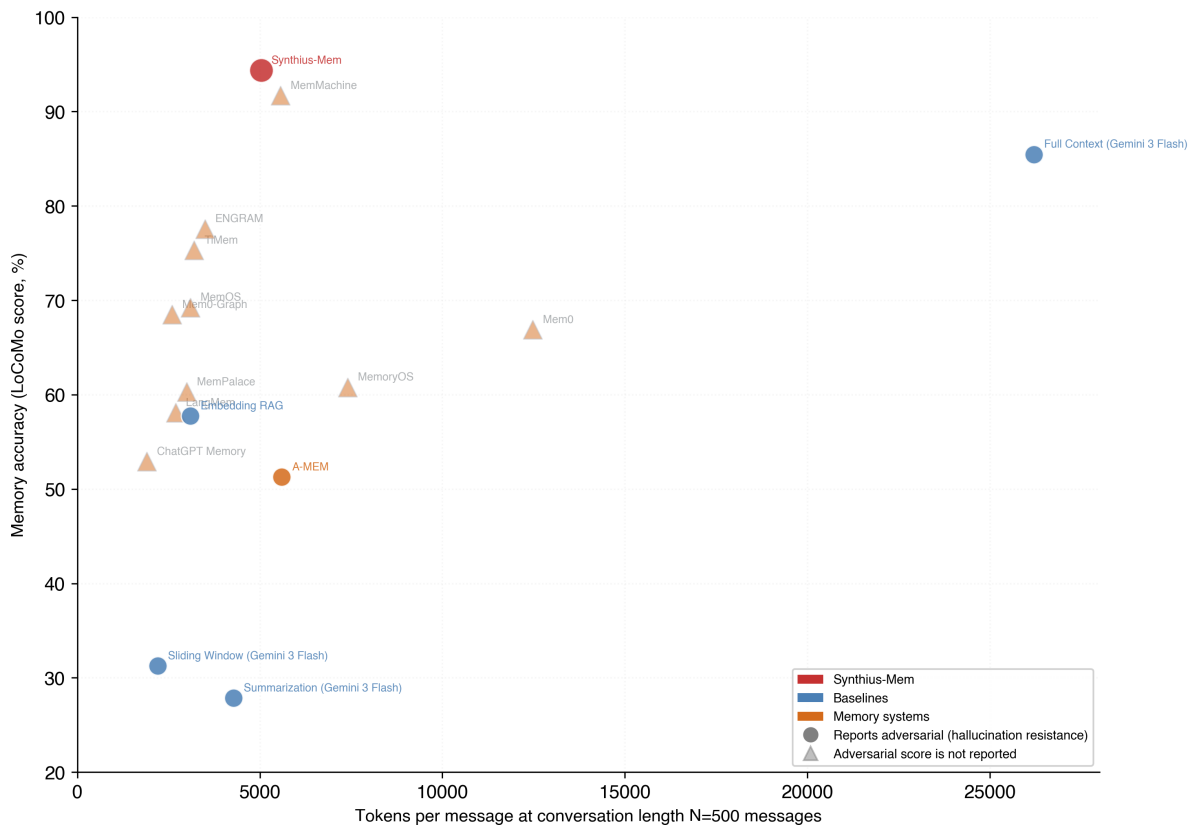

*Figure 5. Token cost vs. accuracy at N=500 (linear scale). Shape: circle = reports adversarial, triangle = adversarial N/R. Color: red = Synthius-Mem, blue = baselines, orange = memory systems. Token estimates in Appendix A.*

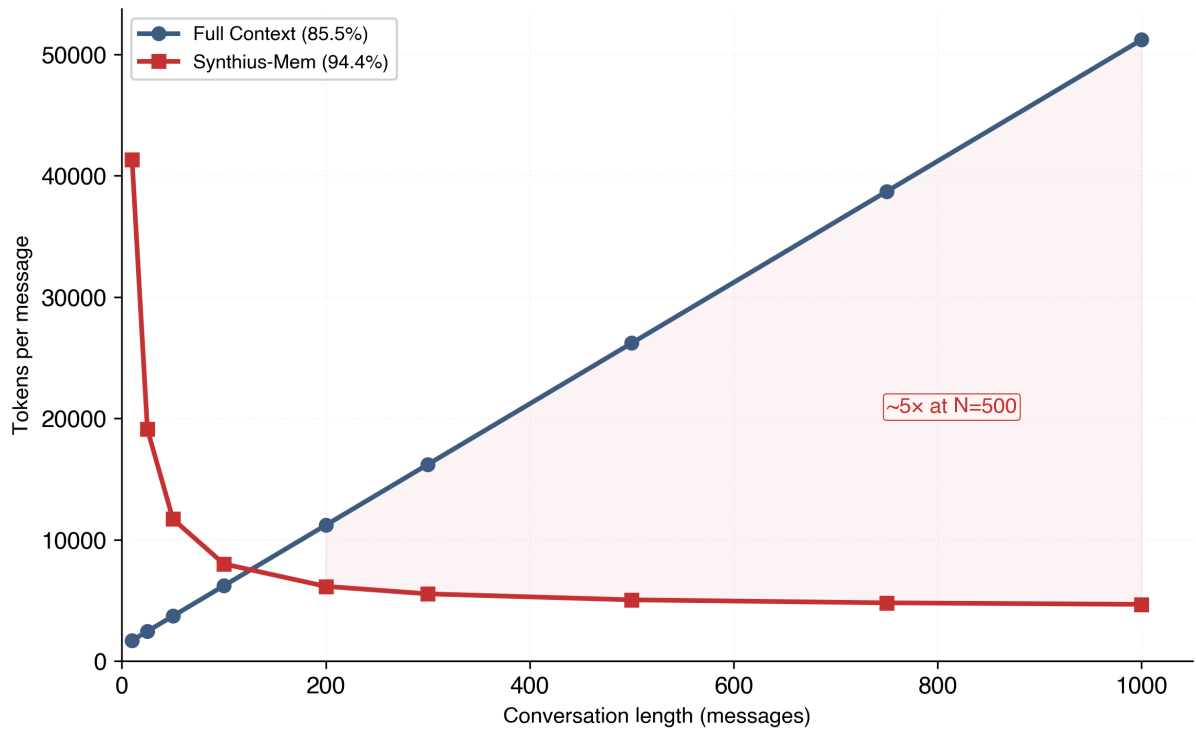

*Figure 6. Token cost scaling with conversation length. Full-context cost grows linearly per message; Synthius-Mem's extraction cost amortizes. At N=500 Synthius-Mem is ~5× more token-efficient while achieving higher accuracy.*

The cumulative token cost over a 500-message conversation is approximately 6.9 million tokens for Full Context versus 4.7 million tokens for Synthius-Mem. The advantage grows with conversation length, since Full Context cost is O(n²) while Synthius-Mem cost is effectively linear after extraction amortizes. Detailed per-system token derivations and the corresponding USD values are provided in Appendix A.

## 4.6 Retrieval Latency

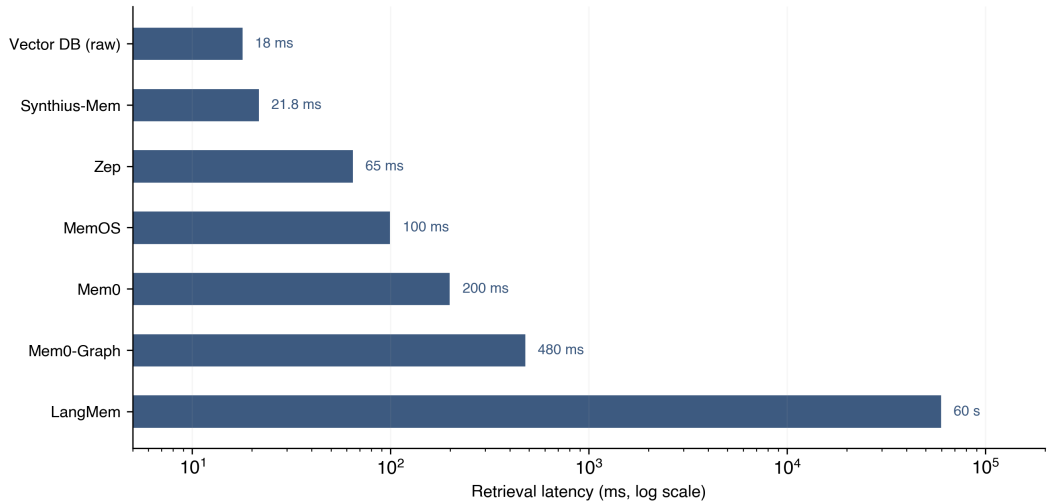

*Figure 7. Retrieval latency for memory systems. Synthius-Mem's CategoryRAG operates at 21.79 ms mean latency.*

Synthius-Mem's CategoryRAG achieves 21.79 ms mean retrieval latency. Figure 9 shows measured latencies for other memory systems where published: Zep (65 ms), MemOS (~100 ms), Mem0 (200 ms), Mem0-Graph (480 ms), LangMem (~60 s). Raw vector database queries (Pinecone, Weaviate) typically achieve 15–20 ms. For conversational agents, retrieval latency is not the user-facing bottleneck: LLM inference time dominates end-to-end response time by two to three orders of magnitude, and the memory retrieval component contributes less than 0.1% of total latency. The practical implication is that CategoryRAG does not degrade user experience regardless of the absolute retrieval speed.

# 5. Discussion

## 5.1 Why Structured Knowledge Retrieval Outperforms Dialogue Retrieval

The retrieval target matters more than the retrieval method. Existing systems retrieve dialogue segments; Synthius-Mem retrieves structured knowledge—pre-parsed facts with metadata, organized in domain-specific schemas. The extraction pipeline performs cognitively demanding inference once, not at every query. The 8.91 pp advantage over full context—despite full context having all information—arises from: (1) higher signal-to-noise ratio, (2) domain-aligned vs. chronological organization, (3) adversarial robustness from extraction filtering (99.55% vs 87.8%), (4) pre-resolved temporal metadata.

Why does a brain-inspired structured approach outperform the main alternatives? Each alternative class sacrifices a different property that biological memory preserves. Full-context replay retains everything but forces the answer LLM to do all the cognitive work—parsing, disambiguation, temporal reasoning, relevance filtering—at every query, which degrades with context length (the "lost in the middle" effect) and scales quadratically in cost. Flat embedding RAG retrieves raw dialogue chunks and suffers from vector space crowding and semantic drift: the embedding of a question about a person's job is similarly close to every mention of the word "job" in the history, producing noisy retrieval that the answer LLM then has to filter. Summarization approaches compress history into LLM-generated narratives but discard exactly the information—temporal ordering, relational nuance, hedged statements—that a persona model needs. Flat fact extraction (Mem0, ChatGPT Memory) does extract structure but treats all facts uniformly: a biographical date, an emotional memory, and a social relationship share the same storage schema, defeating the purpose of extraction in the first place. Ground-truth-preserving systems like MemMachine retain raw episodes with good retrieval but cannot refuse hallucination questions—absence of evidence in their vector store does not signal absence of fact, because every query returns top-K matches regardless of relevance.

The brain does none of these things. Episodic memory binds spatiotemporal context into distinct hippocampal traces. Semantic knowledge lives in hierarchical neocortical categories accessed by cue. Social person-models are maintained by a dedicated mPFC–TPJ network. Evaluative preferences are encoded through dopaminergic reward circuits. Each subsystem has its own encoding scheme, its own consolidation dynamics, and its own retrieval pathway—and together they produce efficient, robust, non-hallucinating memory at twenty watts. Synthius-Mem inherits this organizational principle directly: six typed domains with dedicated schemas, dedicated extraction, dedicated retrieval tools. The structural consequence is that each class of question lands in the right domain with the right metadata, noise is filtered at extraction time rather than query time, and absence of evidence becomes a reliable refusal signal because the knowledge base contains only attested facts. This is the mechanism behind the gap over all three competitor classes on LoCoMo.

## 5.2 The Peripheral Detail Tradeoff

Synthius-Mem's lowest score is on the peripheral detail category. This is intentional. Peripheral details are things like the exact name of a restaurant someone mentioned once, the specific breed label for a pet, the precise adjective used to describe a passing mood, or a symbolic nickname used in jest. These details appear in conversation but do not define who a person is.

If Synthius-Mem stored every such detail, the persona database would grow dramatically. A single 200-message conversation might yield hundreds of peripheral facts—exact phrases, incidental mentions, throwaway remarks. Over months of use with regular uploads, the persona would balloon with noise that would: (1) increase retrieval confusion, since more stored facts mean more candidates during retrieval, increasing the chance of surfacing irrelevant evidence and diluting the signal from important facts; (2) increase storage and processing costs, since each stored fact consumes space, indexing, and tokens when it appears in retrieval context; and (3) not improve user experience, since users do not expect their persona assistant to remember the exact adjective they used in a casual aside—they expect it to know their career, relationships, values, and major life events.

The extraction pipeline therefore deliberately prioritizes biographical facts (name, age, education, family, location), relational structure (who the person knows and how they relate), significant experiences (career moves, life changes, emotional milestones), preferences and values (what matters to the person), and communication style (how they talk, not what they happen to mention). This is a product decision: a persona that perfectly remembers every trivia detail but bloats to unusable size is worse than one that captures the essential character and stays fast, focused, and affordable. The relevance threshold is configurable for applications that genuinely need exhaustive recall (legal transcription, medical records).

## 5.3 Limitations

**Scoring methodology.** Cross-system comparison on LoCoMo is complicated by the fact that different research groups employ substantially different evaluation prompts. The original LoCoMo paper (Maharana et al., 2024) uses token-overlap F1. Mem0 (Singh et al., 2025) instructs its judge to "be generous with grading." Hippocampus uses a 5-point scale. ENGRAM's evaluation prompt is unpublished. Our binary rubric is described in Section 4.1 along with the reasoning for each level. To ensure cross-paper comparability we commonly use; results are in Section 4.2 and demonstrate that our lead is judge-robust. Absolute scores in Table 2 (other systems) should still be interpreted as approximate rankings; the controlled baseline comparison in Table 4—where all systems use identical evaluation conditions—is the most reliable evidence of relative performance.

**Embedding RAG baseline.** Our RAG baseline uses per-session chunking (text-embedding-3-small 1536d, top-3 retrieval), achieving 57.74%. This configuration is representative of realistic deployment but operates on a small corpus (a single conversation). At production scale with thousands of conversations, RAG performance would likely deteriorate further due to vector space crowding and semantic drift (arXiv:2601.15313). Full configuration is in Appendix A.

**Memory system token cost estimates.** Per-message token costs for Mem0, TiMem, MemOS, and other systems shown in Figure 7 are estimates based on comparable LLM retrieval overhead, not measured values. These systems do not publish per-message cost breakdowns. Full derivation methodology and the corresponding USD conversion are in Appendix A.

**Additional benchmarks.** While LoCoMo is the most comprehensive and widely adopted benchmark for conversational memory, it evaluates only one facet of persona memory: factual QA over dialogue history. A complete assessment of Synthius-Mem requires evaluation on complementary benchmarks. Multi-Session Chat (MSC; Xu et al., 2022) would test cross-session persona consistency. CHRONICLE (Chen et al., 2024) would stress-test temporal reasoning beyond what LoCoMo covers. LongMemEval (Wang et al., 2024) would evaluate long-term memory under different criteria. Beyond existing benchmarks, the psychological profiling subsystem requires separate validation against ground-truth assessments from validated psychometric instruments. We discuss the design requirements for a comprehensive persona fidelity benchmark in Section 5.4.

### 5.4 Toward a Comprehensive Persona Memory Benchmark

LoCoMo evaluates factual question-answering over dialogue history—a necessary but insufficient test of persona memory. The field needs a comprehensive benchmark evaluating at least six dimensions:

- Psychological profile accuracy: Validate extracted Big Five, Schwartz Values, and other framework scores against ground-truth assessments from standardized questionnaires.
- Social graph completeness: Entity-level precision and recall for identified persons, relationship-type accuracy, attribute-level errors.
- Preference evolution tracking: Temporal accuracy of preference states, change detection F1, historical consistency.
- Personality consistency in generation: Whether psychometric-grounded responses match the modeled personality.
- Cross-domain coherence: Consistency of facts across domains.
- Memory update fidelity: Accuracy of incremental updates after multiple compaction cycles.

Beyond these dimensions, three structural requirements would substantially improve the integrity of persona memory evaluation. First, the corpus must be at least two orders of magnitude larger than current LoCoMo. Memory systems are designed for conversation histories spanning months or years; benchmarking them on dialogues that span only a handful of sessions measures the wrong regime. A realistic persona memory benchmark should contain conversation histories on the order of tens of thousands of turns per participant—roughly 100× the current LoCoMo scale. Second, the benchmark must include a hidden held-out portion. The visible split is necessary for development, but contamination is inevitable on any public dataset that has been available for a year or more. A held-out split, evaluated only by the benchmark maintainers, prevents both intentional overfitting and accidental leakage through training data. Third, the field needs a single common judge protocol—a fixed prompt, a fixed model, and a fixed scoring rubric—that every paper must use. The current situation where Mem0 uses a generous judge, Hippocampus uses a 5-point scale, ENGRAM uses an unpublished prompt, and the original LoCoMo paper uses F1 makes cross-paper comparison nearly impossible. A standardized judge would reveal which architectural decisions actually matter rather than which evaluation framing flatters the system most.

## 6. Future Work and Vision

We position Synthius-Mem as the memory subsystem of a broader platform for persistent, personalized AI agents. The agent market is projected to reach $52.62B by 2030 (Grand View Research, 2025). Research directions include: real-time streaming extraction, multi-agent shared memory with domain-level access control, temporal decay inspired by Ebbinghaus (1885), active memory acquisition through gap-detection, privacy-preserving memory via federated extraction, and the comprehensive persona benchmark described in Section 5.4. The architecture extends naturally to a Model Context Protocol (MCP) service for memory portability across platforms.

## 7. Conclusion

Synthius-Mem organizes conversational knowledge into six neuroscience-inspired domains. On LoCoMo, it achieves 94.37% weighted accuracy—exceeding TiMem (75.30%) by 19.07 pp and human performance (87.9%) by 6.47 pp. Adversarial robustness reaches 99.55%; core fact accuracy 98.64%; temporal precision 94.40% with zero wrong answers. Even under the strict CategoryRAG retrieval operates at 21.79 ms mean latency. At 500 messages, Synthius-Mem is ~5× more token-efficient than full context while achieving higher accuracy. Structured, domain-specific memory extraction fundamentally outperforms both raw-context and unstructured-memory approaches.

# Appendix A: Cost Models and Per-System Derivation

## A.1 Token-Based Per-Message Cost Model

All systems share a baseline LLM call: ~1,000 token system prompt input + ~200 token output ≈ 1,200 baseline tokens per message. The differential cost depends on the additional input each approach feeds to the answer LLM and any amortized pipeline overhead.

**Table 7. Per-message token cost at N=500**

| Approach | Additional input | Amortized overhead | Total tok/msg |
|---|---|---|---|
| Full Context | 25,000 tok (full history) | — | ~26,200 |
| Synthius-Mem | ~2,000 tok retrieved + 1,100 tok planner | 740 tok extraction | ~5,040 |
| Embedding RAG | ~1,900 tok (3 sessions) | Embedding (negligible) | ~3,100 |
| Summarization | ~3,000 tok (summary + recent) | ~75 tok summarization | ~4,275 |
| Sliding Window | ~1,000 tok (recent msgs) | — | ~2,200 |

## A.2 Cumulative Conversation Token Cost

The cumulative token cost over an N-message conversation is the sum of per-message costs. For Full Context, this grows quadratically because each subsequent message includes a longer history. For Synthius-Mem, the per-message cost decreases as N grows (extraction amortizes), yielding sublinear cumulative growth. At N=500: Full Context cumulative ≈ 6.9M tokens; Synthius-Mem cumulative ≈ 4.7M tokens. At N=1000: Full Context ≈ 27M tokens; Synthius-Mem ≈ 9.3M tokens.

## A.3 USD Cost Model (Secondary)

For readers who prefer monetary units, we provide USD-converted figures. All conversions use GPT-5.4 pricing as of April 2026: $2.50/M input tokens, $15/M output tokens. Note that token cost is the architecture-neutral metric; USD values depend on the chosen model and provider.

**Table 8. Per-message USD cost at N=500 (GPT-5.4 pricing)**

| Approach | Additional input | Amortized overhead | Total USD/msg |
|---|---|---|---|
| Full Context | 25,000 tok | — | $0.068 |
| Synthius-Mem | ~2,000 tok retrieved | $0.31/N extraction | $0.011 |
| Embedding RAG | ~1,900 tok (3 sessions) | Embedding cost | $0.010 |
| Summarization | ~2,500 tok (summary) | Periodic summarization | $0.012 |
| Sliding Window | ~1,000 tok (recent msgs) | — | $0.008 |

Cumulative USD at N=500: Full Context $18.41, Synthius-Mem $7.42 (~2.5× savings). At N=1000: Full Context $68.28, Synthius-Mem $13.57 (~5.0× savings).

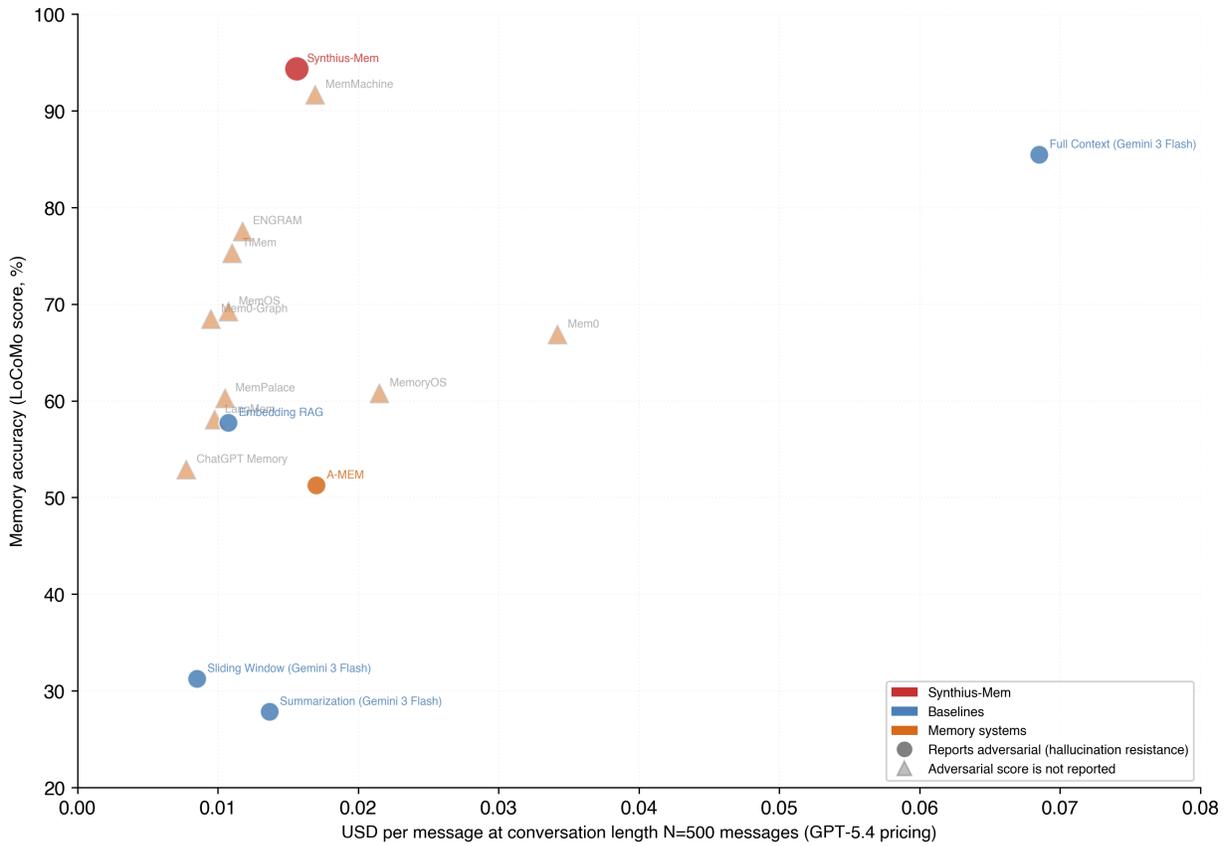

*Figure 8. (Appendix figure) USD cost equivalents of the token-cost analysis from Figure 7. (a) Per-message USD cost at N=500. (b) Cumulative conversation USD cost.*

## A.4 Embedding RAG Baseline Configuration

Our embedding RAG baseline uses per-session chunking with OpenAI text-embedding-3-small (1536 dimensions) and cosine similarity retrieval.

**Table 9. Embedding RAG configuration**

| Parameter | Value |
| --- | --- |
| **Chunk unit** | Entire conversation session (all turns in one session) |
| **Chunks per conversation** | 19–32 |
| **Chunk size** | 264–1,207 tokens (median ~620) |
| **Top-K retrieved** | 3 |
| **Retrieved context size** | ~1.5–3K tokens (3 full sessions) |
| **Embedding model** | text-embedding-3-small (1536 dimensions) |
| **Retrieval method** | Cosine similarity |
| **Accuracy** | 57.74% |

Per-session chunking produces 19–32 chunks per conversation, with top-3 retrieval covering ~10–15% of the conversation content. This is representative of realistic deployment where corpora contain thousands of sessions.

## A.5 Per-System Token-Cost Derivation for Figure 7

Figure 7 plots accuracy against estimated total tokens per message at N=500. Our estimates combine published retrieved-context measurements with a standard overhead model. The methodology is as follows:

For each system, total tokens per message = retrieved context tokens (from published measurements or architecture-based estimates) + 1,000 tokens system prompt + 200 tokens output + additional LLM call overhead (for systems that make multiple LLM calls per query). Where published retrieved-context data is available from third-party measurements, we use it directly. Where only self-reported or no data exists, we estimate from the published architecture and flag the estimate as low-confidence.

**Published retrieved-context measurements**

Four independent sources provide per-query token data for competing systems on LoCoMo:

Source 1: TiMem paper (Li et al., 2026, arXiv:2601.02845, Table 5). Reports retrieved tokens per query on LoCoMo for six systems. This is the broadest multi-system comparison available and forms the primary basis for our estimates.

Source 2: SYNAPSE paper (Jiang et al., 2026, arXiv:2601.02744, Table 4). Reports tokens per query for a partially overlapping set of systems. SYNAPSE measures LangMem at ~717 tok/query and A-MEM at ~2,520 tok/query (consistent with TiMem's 2,431).

Source 3: "Beyond the Context Window" (Pollertlam & Kornsuwannawit, 2026, arXiv:2603.04814). Independent cost analysis measuring Mem0 at 1,046 retrieved tokens per query (top-k=20)—consistent with TiMem's measurement of 1,070.

Source 4: MemMachine paper (Wang et al., 2026, arXiv:2604.04853, Table 8). Reports total pipeline tokens (not just retrieval): MemMachine = 4.20M input tokens across 1,540 questions = ~2,727 tok/query.

Also measures Mem0 main/HEAD at ~12,472 tok/query total pipeline—substantially higher than retrieval-only measurements, highlighting the gap between retrieved context and total system cost.

**Table 10. Published retrieved context tokens per query on LoCoMo (TiMem, Table 5)**

| System | Retrieved tok/query | Source |
| --- | --- | --- |
| **TiMem** | 511 | Self-reported (TiMem paper) |
| **Mem0** | 1,070 | Third-party (TiMem paper) |
| **MemOS** | 1,371 | Third-party (TiMem paper) |
| **A-MEM** | 2,431 | Third-party (TiMem paper) |
| **MemoryOS** | 4,659 | Third-party (TiMem paper) |

**Total token estimates**

We convert retrieved-context measurements to total-token estimates by adding the standard overhead. For systems with multiple LLM calls per query (TiMem: 2 calls for recall planning and gating; A-MEM: up to 13 calls for memory management; MemoryOS: 4.9 calls average as reported in their paper), we add the corresponding overhead. For single-call systems (Mem0, MemOS, ChatGPT Memory), we add only the baseline 1,200 tokens.

**Table 11. Total tokens per message at N=500 (estimates for Figure 7)**

| System | Retrieved tok | Overhead | Total tok/msg | Confidence |
|---|---|---|---|---|
| **Full Context** | 25,000 (all history) | 1,200 baseline | ~26,200 | Calculated |
| **Synthius-Mem** | 2,000 (structured) | 1,200 + 1,100 planner + 740 extraction | ~5,040 | Measured |
| **MemoryOS** | 4,659 (TiMem T5) | 1,200 + ~1,500 (4.9 LLM calls) | ~7,400 | High |
| **A-MEM** | 2,431 (TiMem T5) | 1,200 + ~2,000 (13 LLM calls) | ~5,600 | Medium |
| **MemMachine** | — (agent mode) | — | ~5,564 (total measured) | High |
| **Mem0** | 1,070 (TiMem T5) | 1,200 baseline | ~2,300 | High |
| **Mem0-Graph** | ~1,370 (est.) | 1,200 baseline | ~2,600 | Medium |
| **MemOS** | 1,371 (TiMem T5) | 1,200 + ~500 scheduler | ~3,100 | Medium |
| **TiMem** | 511 (self-reported) | 1,200 + ~1,500 (2 extra calls) | ~3,200 | Medium |
| **LangMem** | 717 (SYNAPSE T4) | 1,200 + ~800 active formation | ~2,700 | Medium |
| **ChatGPT Mem.** | ~500 (est. facts) | 1,200 baseline | ~1,900 | Low |
| **Emb. RAG** | 1,900 (3 sessions) | 1,200 baseline | ~3,100 | Calculated |
| **Summarization** | 3,000 (summary+recent) | 1,200 + 75 amort | ~4,275 | Calculated |
| **Sliding Window** | 1,000 (20 msgs) | 1,200 baseline | ~2,200 | Calculated |

Important caveats. (1) Retrieved-context measurements from the TiMem and SYNAPSE papers reflect only the tokens sent to the answer LLM, not total pipeline overhead including extraction, embedding, and memory management LLM calls. Total tokens are higher. (2) The The MemMachine paper reports two modes: memory mode (4.20M input tokens / 1,540 questions = 2,727 tok/query for bare vector search) and agent mode (8.57M / 1,540 = 5,564 tok/query for LLM-orchestrated retrieval). We use the agent-mode figure (5,564) because it corresponds to the configuration that achieves the reported 91.69% accuracy. The memory-mode figure (2,727) achieves slightly lower accuracy (91.23%) and excludes LLM routing calls. Both figures are total pipeline measurements, making them directly comparable to our Synthius-Mem figure (5,040) but not to the TiMem-table retrieval-only numbers without adjustment. (3) Multi-call overhead estimates (for A-MEM, MemoryOS, TiMem) are derived from call counts reported in the respective papers but the token budgets of individual sub-calls are not always published. (4) ChatGPT Memory token usage is estimated from reverse-engineering reports and may vary by user profile size. (5) No independent audit has measured all systems under identical conditions; the heterogeneity of measurement methodologies is a limitation of the current landscape.

## A.6 Commercial Pricing of Memory Systems

For systems with commercial offerings, we report platform pricing for context. Platform fees reflect infrastructure, storage, and support beyond raw LLM inference and do not directly correspond to the token-cost estimates above.

**Table 12. Commercial memory system pricing (April 2026)**

| System | Pricing Model | Free Tier | Paid Plans |
|---|---|---|---|
| **Mem0** | Per operation | 10K adds + 1K retrieves/mo | $19/mo to $249/mo |
| **Zep** | Per credit | 1,000 credits/mo | $25/mo to $475/mo |
| **Letta** | Subscription + usage | Limited agents | $20–200/mo + LLM costs |
| **LangMem** | Open source | Full (self-hosted) | $0 (infrastructure costs only) |
| **ChatGPT Memory** | Bundled in subscription | N/A | Included in Plus ($20/mo) |
| **TiMem / A-MEM / MemOS** | Research only | Open source | N/A |